# MidNet: An Anchor-and-Angle-Free Detector for Oriented Ship Detection in Aerial Images

Feng Jie, Yuping Liang*, Junpeng Zhang, Xiangrong Zhang, Quanhe Yao, Licheng Jiao

School of Artificial Intelligence, Xidian University, Xi'an, China

jiefeng0109@163.com          yupingliang.nintendo@gmail.com

## Abstract

*Ship detection in aerial images remains an active yet challenging task due to arbitrary object orientation and complex background from a bird's-eye perspective. Most of the existing methods rely on angular prediction or predefined anchor boxes, making these methods highly sensitive to unstable angular regression and excessive hyper-parameter setting. To address these issues, we replace the angular-based object encoding with an anchor-and-angle-free paradigm, and propose a novel detector deploying a center and four midpoints for encoding each oriented object, namely MidNet. MidNet designs a symmetrical deformable convolution customized for enhancing the midpoints of ships, then the center and midpoints for an identical ship are adaptively matched by predicting corresponding centripetal shift and matching radius. Finally, a concise analytical geometry algorithm is proposed to refine the centers and midpoints step-wisely for building precise oriented bounding boxes. On two public ship detection datasets, HRSC2016 and FGSD2021, MidNet outperforms the state-of-the-art detectors by achieving APs of 90.52% and 86.50%. Additionally, MidNet obtains competitive results in the ship detection of DOTA.*

## 1. Introduction

Object detection has been one of the most longstanding and fundamental tasks in computer vision. Among the variety of object detection tasks, ship detection in aerial images has been drawing an increasing attention for territorial defense, fisheries management and marine situational awareness [1, 2]. Due to arbitrary orientation and large aspect ratio of the ship from a bird's-eye view, horizontal bounding boxes are not adequate for describing its location information. In recent years, oriented ship detection has attracted extensive academic attention.

A widely-adopted paradigm for handling oriented objects is to attach an extra unit to the boundary box regression branch for predicting rotation angles. Fig. 1 (a) and (b) illustrates two basic coding manners for angular prediction in anchor-based and anchor-free forms. Although there will be varieties in the details of coding manners, the angular-based paradigm limits their further improvement because of the following two reasons: 1) *The boundary discontinuity of regression*: due to the periodicity of angles and the exchangeability of edges, a small prediction deviation may bring a great loss increase under the boundary condition, such as approximately vertical or horizontal objects. 2) *High sensitivity to angular prediction errors*: A slight angular prediction error will cause huge Intersection of Union (IoU) reductions, especially for objects with large areas or large aspect ratios. These both make the learning unstable with even slight angular deviations.

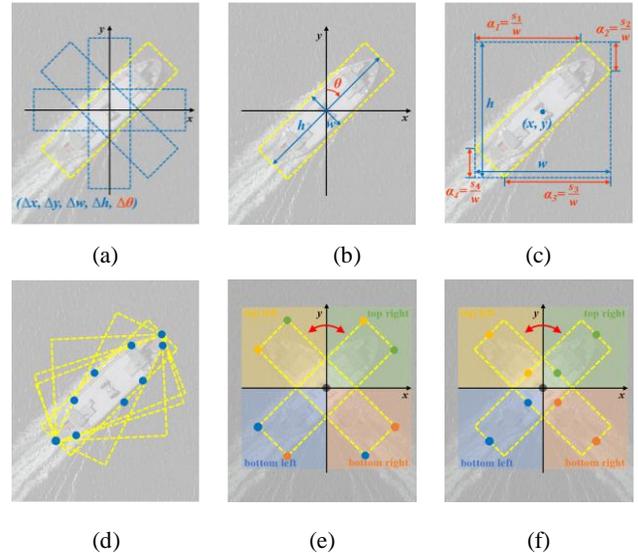

Figure 1. The coding manners of oriented object detection. (a) and (b) show two encoding manners based on angle prediction, which are implemented in anchor-based and anchor-free form respectively. (c) shows an angle-free prediction method, but this method still has the limitation of anchor mechanism. (d)-(f) are keypoint detectors. (d) shows RepPoint case, where the yellow represents the potential bounding boxes, and these bounding boxes are not unique for an oriented object. (e) shows corner case. As the angle changes, the top left corner may fall in the top left or top right quadrant. The orientation definition of corners is ambiguous for oriented objects. (d) shows the proposed midpoint case. No matter how the angle is rotated, the orientation of the midpoint corresponds to that of the quadrant. It is beneficial to stable learning of features.

To alleviate unstable angle learning, there are different attempts to transform angular prediction into other ways. Some researchers locate the intersection of the object and its external rectangle to rotate a horizontal bounding box into an oriented one. [3] regresses four length ratios representing the relative gliding offset on each corresponding side, shown in Fig. 1(c). [4] predicts the offsets from the middle position of the sides to the intersection. These methods require predefined horizontal boxes as the starting status, which is usually implemented in the form of anchors. Thus, they inevitably cause excessive manually adjusted hyper-parameters, which limits the generalization capability for novel scenarios, and consumes excessive computational cost.

Keypoint-based detector [5-8] has attracted extensive academic attention in horizontal object detection by providing an alternative method for encoding the object location using representative keypoints. Concurrently, it provides a new perspective for angle transformation in oriented object detection. However, *the definition of keypoints in most existing detectors is confusing for oriented objects*. Take the corner-based detectors as an example, the corners can always be defined by their relative positions to the center point in the horizontal box. However, for the oriented object, the "top left" corner may appear in the top left or top right quadrant due to the arbitrary orientation (Fig.1 (e)), which will not only make the definition for corners ambiguous, but also cause instability in the process of feature learning. Another challenge in using keypoints for encoding rotated objects lies on *the uncertainty of the side direction of objects*. For the horizontal object, there are always two vertical and horizontal sides in the bounding box. Therefore, a unique external bounding box can be construct according to the keypoint estimation, such as RepPoint [5], ExtremeNet [6]. On the contrary, for the oriented object, the uncertainty of the side direction makes the external rectangle of the keypoints non-unique, as shown in Fig. 1 (d). Thus, it is difficult to build a precise bounding box without angle information. In [1], Zhang *et al.* proposed an oriented ship detector, which locates the object through the head and center point. However, this method requires additional priori information of the head points and increases the cost of manual annotation.

In this paper, we propose a keypoint-based representation for oriented ship detection, and design a novel anchor-and-angle-free detector, MidNet. The midpoints of the four edges and the center point of the object are utilized to encode the bounding box, as shown in Fig. 1(f). Compared with corner points, the midpoints are more likely to fall in the object, which makes them contain more abundant features. Moreover, the direction of the midpoints of the four edges always corresponds to that of the quadrant (top left, top right, bottom right and bottom left). The midpoints-based representation enables the network to exploit the information inside the object in a more definite way and extract discriminative features for learning.

According to this idea, MidNet predicts the center point and its maximum inscribed circle radius, as well as the midpoints and their centripetal vectors. For the midpoint, a novel deformable convolution is designed to enhance the features by using the priori information of ship's structural symmetry, especially for trailing or obscured ships. After obtaining the keypoints estimation, the keypoints are suppressed and matched adaptively according to the learned radiuses and centripetal shifts, and the accurate bounding box is built by the refined keypoints through an analytical geometry method.

The contributions of this paper can be summarized as follows:
- A novel oriented object representation based on midpoints is proposed, which not only makes the framework avert the instable angular regression and excessive hyper-parameter adjustment, but also eliminates the confusion in the keypoint definition and learning process.
- An anchor-and-angle-free detector MidNet is proposed, where a symmetrical deformable convolution is customized for feature learning of ship objects. Moreover, a purely analytical geometry method is designed to refine and construct keypoints into precise oriented rectangular boxes without any additonal information.
- Extensive experiments and visual analysis on different datasets and detectors prove the efficacy of our technique. The proposed method achieves 90.52% and 86.50% APs on HRSC2016 and FGSD2021, which outperforms the state-of-the-art methods. Moreover, it achieves 86.42% AP for the ship detection on DOTA dataset.

## 2. Related Work

### 2.1. Oriented Object Detectors

Arbitrary-oriented detectors are widely used in remote sensing and scene text images. Most of the early methods derived from the anchor-based detectors. RRPN [9] obtains rotated region proposal by tiling rotated anchors. RoI-Trans [10] transformed a horizontal RoI into a rotated RoI (RRoI) through RRoI leaner and RRoI wraping. In R$^3$Det [11], a refined single-stage rotated detector was proposed for the feature misalignment problem. Like the horizontal detectors, these anchor-based methods suffer from excessive hyper-parameters, high computational cost and low adaptability.

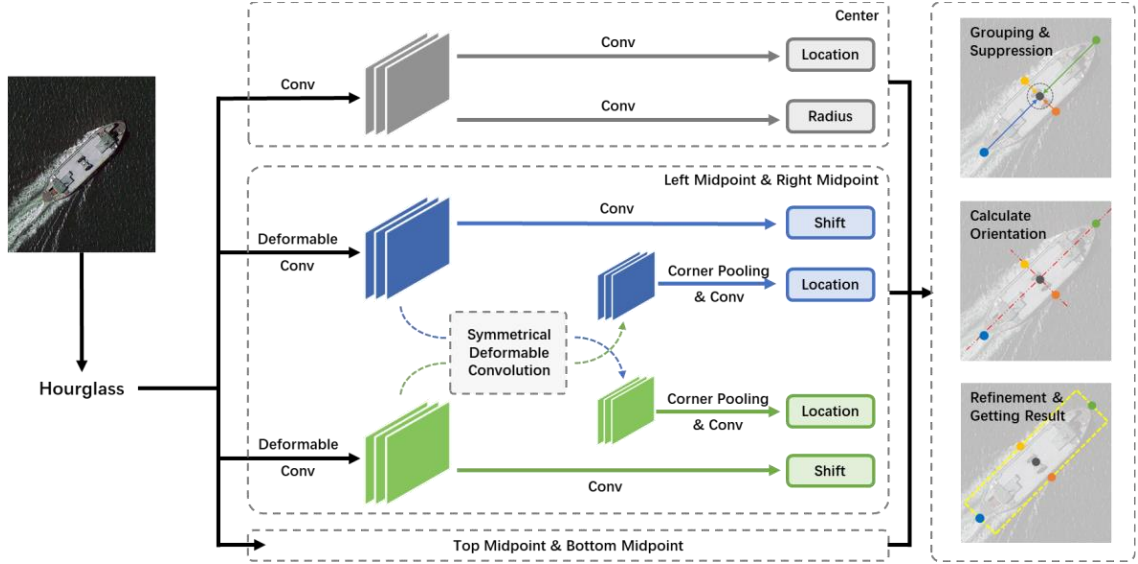

Figure 2. The overall framework of MidNet. MidNet predicts five keypoints (one center and four midpoints) for ascribing an oriented boundary box. For a pair of symmetric midpoints, such as top left and bottom left midpoints, MidNet predicts the centripetal shift and jointly learns the enhanced features by the symmetrical deformable convolution. For center, its matching radius is also predicted. Combining the keypoint, radius, centripetal shift and confidence score, a series of analytical geometry post-processing is used to group, suppress, and refine the keypoints. Finally, the keypoints are transformed into oriented bounding boxes.

Meanwhile, the unstable angle prediction is another central issue in oriented object detection. In CSL [12] and DCL [13], angle regression was converted into a classification task to handle the boundary problem. In Gliding Vertex [3] and Oriented R-CNN [4], the frameworks construct the oriented bounding box by locating the intersection of the object and its external rectangle. However, these methods either fail to completely avoid angle prediction, or require predefined anchors, which limits the further improvement of performance.

### 2.2. Keypoint-Based Detectors

For the horizontal object detection, anchor-based detectors suffer from sample imbalance, redundant hyper-parameters and limited generalization ability. The keypoint-based methods have attracted extensive academic attention. These approaches generate the bounding boxes by a set of keypoints belonging to the objects. CornerNet [7] predicts the top-left and bottom-right corners, and matches them by calculating the distance of the embedded vectors. To enriching information at the central regions, CenterNet [8] predicts the center points additionally on the original structure. However, for objects with similar appearance features, the above two methods may produce similar embedding vectors, resulting in incorrect keypoint matching. CentripetalNet [14] predicts the position and the centripetal shift of the corner points, and matches the corners whose shifted results are aligned. Except for corners, ExtremeNet uses extreme points on the four edges to represent the object, but this method needs more accurate annotation. Different from ExtremeNet, RepPoint adopts a top-down fashion and utilizes deformable convolution to locate the representative points of the object.

A series of keypoint-based methods have made remarkable achievements. However, due to the challenges of extracting rotation features and encoding the oriented bounding box, these methods have received little attention in the oriented object detection.

### 2.3. Oriented Ship Detectors

Ship detection receives additional attention in the remote sensing data analysis due to the unique importance in rescue and military. [15] developed the HRSC2016 dataset in 2017, which is one of the most important benchmarks in this field. As early attempts, [16, 17] focused on improving the horizontal RPN and ROI pooling to obtain high-quality rotation proposals. However, considering the limitations of anchor mechanism, some researchers focused on anchor-free method in recent year. In order to accurately predict the position and size of the oriented objects, [2] proposed a rotation Gaussian-Mask model, and [18] proposed feature extraction module (FEM). These methods modified particular modules on the original paradigm to adapt rotation feature learning, rather than make further exploration in the customized representation for oriented ships.

## 3. MidNet

MidNet predicts five keypoints $\{p_l, p_t, p_r, p_b, p_c\}$ to represent an object, where $\{p_l, p_t, p_r, p_b\}$ are the midpoints on the four sides, and $p_c$ refers to the center point of the object. We define the point located in the top left quadrant or directly left to the center point as the left midpoint $p_l$, and define the others ($p_t, p_r, p_b$) in clockwise order. Among four midpoints, a pair of symmetric midpoints form a group. As shown in Fig. 2, MidNet contains three branches, one of which is used to predict the center and its matching radius, and the other two branches of the same structure predict the midpoint groups and centripetal shifts. In the midpoint branch, a symmetrical deformable convolution is designed to promote the feature learning of symmetrical midpoints. After obtaining the keypoint estimation, all the midpoints and centers are grouped and suppressed according to the matching radius and centripetal shift. Then, the orientation of the object is precisely calculated under the criterion of minimum confidence-sensitive offset, and the position of the keypoints is refined according to the axisymmetric characteristics of ships. Finally, the refined keypoints are constructed accurate oriented bounding boxes.

### 3.1. Keypoint Estimation

**Midpoint Prediction.** In the midpoint branch, a pair of symmetric midpoints are jointly learned. For each midpoint, it is described by its coordinate, confidence score and centripetal shift, which are used for positioning, classification and grouping, respectively. Inspired by CentripetalNet, the centripetal shift is defined by the relative position of a midpoint to the center:

$$cs_i = (|x_i - x_c|, |x_i - y_c|), \quad \forall i \in \{l,t,r,b\} \quad (1)$$

where the $(x_i, y_i)$ denotes the coordinates of the midpoint in four directions, and $(x_c, y_c)$ refers to the coordinates of the center. In a pair of symmetric midpoints, a new deformable convolution is proposed to predict the location of midpoints more accurately. It should be noted that benefit from the generality of the midpoint distribution, the sign of the centripetal shift is fixed for each midpoint. Thus, only the value of centripetal shift needs to be predicted, which eases the regression process.

In the training phase, the coordinates and confidence scores of midpoints are obtained by the CornerNet pipeline, which predicts a heatmap for locating the midpoints. The focal loss is calculated based on the rendered Gaussian map. In order to simplify the structure, we omit the offset branch in the CornerNet, and refine the position of keypoints in the post-processing process in a simple way. For midpoints in different directions, the symmetrical deformable convolution and corresponding corner pooling are used to aggregate the internal features of the object. The L1 loss is used for the centripetal shift. For midpoints, the centripetal shift prediction is category-agnostic, but the heatmap specific.

**Center Point Prediction.** In aerial images, the sizes of ships range from 300×300 pixels (aircraft carriers) to 10×10 pixels (yachts). Meanwhile, many ships are densely arranged side by side in the ports. Due to inevitable shift prediction error and inflexible manually-set matching range, some existing keypoint matching methods, such as CentripetalNet and CenterNet, would cause incorrect keypoint matching for ship detection. In MidNet, the center point and its matching radius are incorporated to group the midpoints adaptively, and facilitate the learning of the information inside the object. The center point is described by the position and matching radius, where the matching radius $r$ is defined by the maximum inscribed circle radius and calculated by the distance from the center point to the nearest midpoint:

$$r = \min\left(\left\|(x_c - x_i, y_c - y_i)\right\|_2, \forall i \in \{l,t,r,b\}\right) \quad (2)$$

In the training phase, the matching radius is supervised by L1 loss, and the location by the focal loss.

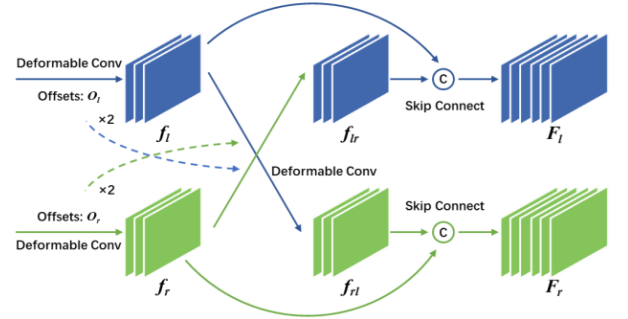

Figure 3. The structure of symmetrical deformable convolution.

### 3.2. Symmetrical Deformable Convolution

The superiority of MidNet relies on accurate locating of keypoints. However, the keypoints of ships are often disturbed in a complex port background or the tailed waves in the water, which causes the difficulties in feature learning of keypoints. To conquer this issue, a novel symmetrical deformable convolution is designed to exploit the features of each other through the mutually symmetrical midpoints. In MidNet, the midpoint branch is responsible for the prediction of a pair of symmetric midpoints. Generally, the feature maps from the backbone is expected to contain salient features of the center regions of the object, and the feature maps in midpoint branch has a significant response to midpoints. Due to the structural characteristics of the ship, the offset from the midpoint to the center is half of that from the midpoint to the symmetrical one. Based on this observation, two deformable convolutions from the same midpoint branch can share the offset to realize the feature enhancement and mutual promotion of symmetric midpoint learning. Taking $p_l$ and $p_r$ as an example (as shown in Fig. 3), the $O_l$ is assumed to be ideal offset from the center to the $p_l$, thus, twice $O_l$ should equal to the offset from $p_l$ to $p_r$. Therefore, by implementing the deformable convolution with $2 \cdot O_l$ on the $f_l$, the $f_{lr}$ could perceive the feature around $p_r$.

### 3.3. Build Precise Oriented Bounding Box.

**Adaptive Keypoint Grouping and Suppressing Scheme.** The estimated keypoints are grouped and suppressed and according to the shifted positions and confidence scores. For each midpoint, its shifted position is calculated by applying its corresponding shifts. A midpoint is selected if its shifted position falls within the matching region defined by the closet center and its associated matching radius, and a center is considered valid when at least one midpoint from each direction matches it. In the cases, where a set of midpoints in the same direction falls in a valid center, only the one with the highest confidence score will be selected while the others are suppressed. Compared with manually set matching schemes in CentripetalNet and ExtremeNet, the proposed grouping scheme can adjust the matching region adaptively according to different sized targets, which improves the detection performance for crowded and size-various ships.

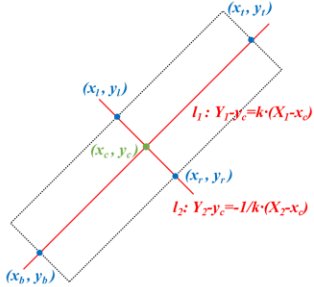

Figure 4. The oriented calculation of the bounding box.

**Oriented Calculation and Keypoint refinement.** For keypoint-based detectors, the prediction of keypoints is easily prone to deviations due to the interference of complex backgrounds, especially for busy ports. A pure analytic geometry method is proposed to refine the keypoints and construct oriented bounding boxes under the criterion of minimum confidence-sensitive offset for keypoints. Firstly, the center is refined by weighted-averaging the coordinates of itself and the shifted positions of its grouped midpoints:

$$x_c^* = \sum_{i \in \{c,l,t,r,b\}} \frac{w_i \cdot \hat{x}_i{'}}{5}, \quad y_c^* = \sum_{i \in \{c,l,t,r,b\}} \frac{w_i \cdot \hat{y}_i{'}}{5} \quad (3)$$

where $\hat{x}_i{'}$ and $\hat{y}_i{'}$ refers to predicted keypoint shifted location, $w_i = \hat{s}_i / \sum_{j \in \{c,l,t,r,b\}} \hat{s}_j$ refers to the normalized confidence weight for the center and the shifted midpoints $i \in \{c,l,t,r,b\}$. Based on the refined centers, the direction of the rotation bounding box with the minimum deviation from the predicted midpoints is formulated and derived. Fig. 4 illustrates the oriented calculation of the bounding box. According to the geometric characteristics of the ideal rectangle, the connecting lines of symmetric midpoints form two symmetric axes of the rectangle. These two symmetric axes intersect the center and are perpendicular to each other. Thus, the axes can be expressed as:

$$L_1 : Y_1 - y_c^* = k(X_1 - x_c^*),$$
$$L_2 : Y_2 - y_c^* = -\frac{1}{k}(X_2 - x_c^*) \quad (4)$$

The deviation distance from the midpoint to its closest symmetric axis can be obtained:

$$d_i = \frac{|k\hat{x}_i{'} - \hat{y}_i{'} - kx_c^* + y_c^*|}{\sqrt{k^2 + 1}}, \quad \forall i \in \{l, r\},$$

$$d_i = \frac{|-\frac{1}{k}\hat{x}_i{'} - \hat{y}_i{'} + \frac{1}{k}x_c^* + y_c^*|}{\sqrt{\frac{1}{k^2} + 1}}, \quad \forall i \in \{t, b\} \quad (5)$$

To make the rotation direction consistent with the keypoint prediction as much as possible, the gradient $k^*$ can be solved by minimizing the summed deviation distances:

$$k^* = \arg\min_k (d_l + d_t + d_b + d_r) \quad (6)$$

The orientation of the rotation object can be obtained from the gradient $k^*$ by only using keypoint matching and refinement. We provide an implicit expression of the angle evaluation by transforming the keypoints, which avoids unstable angle prediction effectively. Finally, project the midpoint on its nearest symmetric axis, calculate the distance from the midpoint to the center, and refine the midpoint along the symmetric axis based on the confidence-weighted. In this case, a unique oriented rectangular box can be obtained according the keypoints, and its score is the average confidence of the five keypoints.

## 4. Experiment

To verify the effectiveness of MidNet, extensive experiments are implemented based on HRSC2016, FGSD2021 and DOTA datasets.

Table 1. Detection accuracy on the HRSC2016 dataset.

| Type | Method | Backbone | Input_Size | mAP |
|---|---|---|---|---|
| Two-stage | RRPN [9] | ResNet-101 | 800×800 | 79.08 |
| | RC2 [16] | - | - | 75.7 |
| | RoI-Trans. [10] | ResNet-101 | 512×800 | 86.2 |
| | Gliding Vertex [3] | ResNet-101 | 512×800 | 88.2 |
| | CSL [13] | ResNet-50 | 800×800 | 89.62 |
| | ReDet [20] | ResNet-101 | 512×800 | 90.46 |
| | Oriented R-CNN [4] | ResNet-101 | 1333×800 | 90.5 |
| One-stage | RSDet [22] | ResNet-50 | 800×800 | 86.5 |
| | DAL [23] | ResNet-101 | 416×416 | 88.95 |
| | RetinaNet-R | ResNet-101 | 800×800 | 89.18 |
| | R$^3$Det [11] | ResNet-101 | 800×800 | 89.26 |
| | DCL [12] | ResNet-101 | 800×800 | 89.46 |
| | CFC-Net [24] | ResNet-101 | 800×800 | 89.7 |
| | GWD [25] | ResNet-101 | 800×800 | 89.85 |
| | S$^2$ANet [26] | ResNet-101 | 512×800 | 90.17 |
| | AR$^2$Det [18] | ResNet-34 | 512×512 | 89.58 |
| | GRS-Det [2] | ResNet-101 | 800×800 | 89.57 |
| One-stage | MidNet (Our) | Hourglass104 | 512×512 | **90.52** |

## 4.1. Datasets

**HRSC2016.** HRSC2016 is a challenging remote sensing ship detection dataset, which collects 1061 images with 2976 instances from six famous ports on Google Earth. The image size of this dataset ranges from 300 × 300 to 1500 × 900. This dataset includes 436, 181 and 444 images for training, validation and test respectively. We use the training set in the training phase and evaluate the detection performance in the test set.

**FGSD2021.** FGSD2021 is another high-resolution satellite ship detection dataset, which is collected from the open Google Earth. The dataset contains 636 images, 5274 instances with a standardized GSD, and 1 meter per pixel. In FGSD2021, the width of the image is ranged from 157 to 7789 pixels, and the height is from 224 to 6506 pixels. Compared with HRSC2016, FGSD2021 has much larger image size. The FGSD2021 dataset is divided into 424 training images and 212 test images. Following the literature [1], all the 20 annotated categories are used to evaluate the model.

**DOTA.** DOTA [19] contains 2,806 aerial images from different sensors and platforms, and the size of the image ranges from around 800 × 800 to 4000 × 4000 pixels. Half of the original images are randomly selected as the training set, 1/6 as the validation set, and 1/3 as the testing set. In order to be consistent with the experimental setting of compared methods, all the categories are used for training and testing. In this paper, we only focus on the accuracy of the ship. The experimental results are obtained by DOTA evaluation server.

## 4.2. Experimental Setting

**Multi-task Training.** The final loss function is:
$$L = L_m + \alpha L_c + L_{cs} + \beta L_r \quad (7)$$
where $L_m$ and $L_c$ are the focal loss for the midpoints and the centers, $L_{cs}$ is the L1 loss for the centripetal shift of four midpoints, and $L_r$ is the L1 loss for the matching radius of the center. Since the number of midpoints in the objects are four times than that of the centers, we set $\beta$ to 0.25. To emphasize the features of the central area of the object, we will set $\alpha$ to 0.5. As in CornerNet, we add intermediate supervision when we use Hourglass-104 as the backbone network.

**Implementation details.** We implemented our method in PyTorch 1.8 and Ubuntu 18.04 with 2 NVIDIA RTX 3090 GPUs. Hourglass [27] is used as the backbone and pretrained on the COCO [21]. We trained 175 epochs with 4 batch sizes in the training phase. For the SGD, the initial learning rate is 0.0001, and it decays 10 times at the 50, 150 epochs. For HRSC2016 dataset, the shorter sides are resized to 512 without changing the image scale. For FGSD2021 and DOTA datasets, we cropped the images into 768×768 with an overlap of 200 pixels. It is worth noting that no more data augmentation is implemented except for random horizontal flip and vertical flip on all the three datasets.

## 4.3. Comparison with State-of-the-Arts

We compare our proposed method with the State-of-the-Art methods on the three datasets, and Tables 1~3 report the detailed comparison results. All the methods are evaluated under Pascal VOC 2007 metrics. For fair comparison, CHPDet [1] is not considered as a comparison because it introduces additional annotation information.

For the HRSC2016 dataset, Oriented R-CNN and S$^2$ANet achieved the best result in single-stage and two-stage methods with the AP of 90.50% and 90.17% respectively. Among anchor-free methods, GRS-Det achieves the highest 89.57% AP. Our proposed method achieves 90.52% AP, and outperforms all the existing detectors.

For the FGSD2012 dataset, 20 categories are used to evaluate the detectors. MidNet achieves an AP of 86.5%, and improves by 1.04% compared with the state-of-the-art method. This proves that the MidNet also has advantages in the multi-classification task.

For DOTA dataset, our method achieves 86.42% AP, which outperforms most single-stage algorithms. And it is competitive with the state-of-the-art two-stage method (Oriented R-CNN). This verifies the effectiveness of our framework.

Table 1. Detection accuracy on different types of ships and overall performance with the state-of-the-art methods on FGSD2021. The short names for categories are defined as (abbreviation-full name): Air - Aircraft carriers, Was - Wasp class, Tar - Tarawaclass, Aus - Austin class, Whi - Whidbey Island class, San -San Antonio class, New - Newport class, Tic - Ticonderoga class, Bur- Arleigh Burke class, Per - Perry class, Lew -Lewis and Clark class, Sup - Supply class, Kai - Henry J. Kaiser class, Hop- Bob Hope Class, Mer - Mercy class, Fre - Freedom class, Ind - Independence class, Ave - Avenger class, Sub – Submarine and Oth - Other.

| Method | Air | Was | Tar | Aus | Whi | San | New | Tic | Bur | Per | Lew | Sup | Kai | Hop | Mer | Fre | Ind | Ave | Sub | Oth | mAP |
|---|---|---|---|---|---|---|---|---|---|---|---|---|---|---|---|---|---|---|---|---|---|
| CSL [19] | 89.7 | 81.3 | 77.2 | 80.2 | 71.4 | 77.2 | 52.7 | 87.7 | 87.7 | 74.2 | 57.1 | **97.2** | 77.6 | 80.5 | **100** | 72.7 | **100** | 32.6 | 37 | 40.7 | 73.73 |
| DARDet [28] | 90.9 | 89.2 | 69.7 | 89.6 | 88 | 81.4 | 90.3 | 89.5 | 90.5 | 79.7 | 62.5 | 87.9 | 90.2 | 89.2 | **100** | 68.9 | 81.8 | 66.3 | 44.3 | 56.2 | 80.31 |
| DCL [12] | 89.9 | 81.4 | 78.6 | 80.7 | 78 | 87.9 | 49.8 | 78.7 | 87.2 | 76.1 | 60.6 | 76.9 | 90.4 | 80 | 78.8 | 77.9 | **100** | 37.1 | 31.2 | 45.6 | 73.34 |
| Oriented R-CNN [4] | 90.9 | 89.7 | 81.5 | 81.1 | 79.6 | 88.2 | **98.9** | 89.8 | 90.6 | 87.8 | 60.4 | 73.9 | 81.8 | 86.7 | **100** | 60 | 100 | 79.4 | 66.9 | 63.7 | 82.54 |
| R³Det [11] | 90.9 | 80.9 | 81.5 | 90.1 | 79.3 | 87.5 | 29.5 | 77.4 | 89.4 | 69.7 | 59.9 | 67.3 | 80.7 | 76.8 | 72.7 | 83.3 | 90.9 | 38.4 | 23.1 | 40 | 70.47 |
| ReDet [20] | 90.9 | 90.6 | 80.3 | 81.5 | **89.3** | 88.4 | 81.8 | 88.8 | 90.3 | **90.5** | 78.1 | 76 | **90.7** | 87 | 98.2 | 84.4 | 90.9 | 74.6 | **85.3** | 71.2 | 85.44 |
| Retinanet-Rbb | 89.7 | 89.2 | 78.2 | 87.3 | 77 | 86.9 | 62.7 | 81.5 | 83.3 | 70.6 | 46.8 | 69.9 | 80.2 | 83.1 | **100** | 80.6 | 89.7 | 61.5 | 42.5 | 9.1 | 73.49 |
| ROI-Trans. [10] | 90.9 | 88.6 | **87.2** | 89.5 | 78.5 | 88.8 | 81.8 | 89.6 | 89.8 | 90.4 | 71.7 | 74.7 | 73.7 | 81.6 | 78.6 | **100** | 75.6 | 78.4 | 68 | 66.9 | 83.48 |
| RSDet [22] | 89.8 | 80.4 | 75.8 | 77.3 | 78.6 | 88.8 | 26.1 | 84.7 | 87.6 | 75.2 | 55.1 | 74.4 | 89.7 | 89.3 | **100** | 86.4 | **100** | 27.6 | 37.6 | 50.6 | 73.74 |
| S²ANet [28] | 90.9 | 81.4 | 73.3 | 89.1 | 80.9 | 89.9 | 81.2 | 89.2 | **90.7** | 88.9 | 60.5 | 75.9 | 81.6 | 89.2 | **100** | 68.6 | 90.9 | 61.3 | 55.7 | 64.7 | 80.19 |
| SCRDet [29] | 77.3 | 90.4 | 87.4 | 89.8 | 78.8 | **90.9** | 54.5 | 88.3 | 89.6 | 74.9 | 68.4 | 59.2 | 90.4 | 77.2 | 81.8 | 73.9 | **100** | 43.9 | 43.8 | 57.1 | 75.90 |
| MidNet (Our) | 90.9 | 88.6 | 79 | 89 | 83.2 | 89.2 | **98.9** | 88.9 | 90.3 | 90.1 | 62.5 | 82.4 | 89.8 | 95.8 | 95.5 | 87.6 | 87 | **84.5** | 83.9 | 73.1 | **86.50** |

Table 3. Ship detection accuracy on the DOTA dataset.

| Type | Method | Backbone | mAP |
|---|---|---|---|
| Two-stage | Faster R-CNN | R-50-FPN | 77.11 |
| | RoI Transformer [9] | ResNet-101 | 83.59 |
| | SCRDet [29] | ResNet-101 | 72.41 |
| | CenterMap-Net [30] | ResNet-50 | 78.1 |
| | Oriented RCNN [4] | ResNet-101 | **87.52** |
| One-stage | RetinaNet-O | ResNet-50 | 79.11 |
| | DRN [31] | Hourglass-104 | 84.94 |
| | PIoU [32] | DLA-34 | 64.8 |
| | DAL [23] | ResNet-50 | 79.74 |
| | RSDet [22] | R-101-FPN | 72.41 |
| | MidNet (Our) | Hourglass-104 | **86.42** |

The visual detection results of MidNet are shown in Fig. 5. It can be seen that MidNet can maintain its advantages for the ships with large aspect ratios or tailing ships in the complex background and crowded scenes.

### 4.4. Ablation Studies

To check the effectiveness of each part in MidNet, the ablation experiment is implemented. Table 4 verifies each module in the network, and Table 5 investigates the effectiveness of the pro-processing method.

**Corners vs. midpoints.** To verify the superiority of midpoint representation, we use MidNet to predict the corners for comparison in Table 4. Due to the uncertainty of corner distribution, we remove the corner pooling to avoid chaotic feature learning. Experiments show that without any additional structure, midpoint-based feature learning improves the AP by 2.23% on the HRSC2016 dataset.

Table 4. Ablation experiments for each module in MidNet on HRSC2016 dataset.

| Corners | Midpoints | Center Prediction | Symmetrical DCN | mAP |
|---|---|---|---|---|
| √ | | | | 86.24 |
| | √ | | | 88.87 |
| | √ | √ | | 89.65 |
| | √ | √ | √ | **90.52** |

**Center Point Prediction.** In the aerial images, the size of ships varies greatly, and there are many long and narrow ships. If the manually-set matching range is too large, it may cause incorrect matching of keypoints with low confidence. If the range is too small, it may cause missed detection. MidNet adaptively delimit the matching range by predicting the center point and its matching radius. Experiment shows that the midpoint-based model improved the AP by 0.78% through incorporating the center prediction.

**Symmetrical Deformable Convolution.** The experimental results show that the detection accuracy is improved by 0.87% by using effective symmetrical deformable convolution. By sharing the offset, the optimization of the feature map and offset promotes each other, which is the key for MidNet to locate the keypoints accurately.

**Post-processing method.** To verify the effectiveness of proposed keypoint matching strategy, we implement the ExtremeNet pipeline on MidNet. In ExtremeNet, any valid combination of keypoints is retained to build a bounding box, and then NMS is carried out in the whole image. As shown in Table 5, this exhaustive strategy reduces the AP by 0.48%, which is caused by high-confidence incorrect keypoint matchings. In addition, the time complexity of this method reaches $O(n^4)$, which is unacceptable in the crowded remote sensing scene. Moreover, a simplified algorithm for bounding box construction without gradient calculation is designed for comparison. Specifically, connect the symmetrical midpoints, and then translate the connection until it passes through the remaining two midpoints, and get the two sides of the bounding box. Repeat this strategy at another pair of midpoints to get the remaining two sides, so the bounding box is generated. Without angle calculation and keypoint refinement, the detection accuracy of the model decreases.

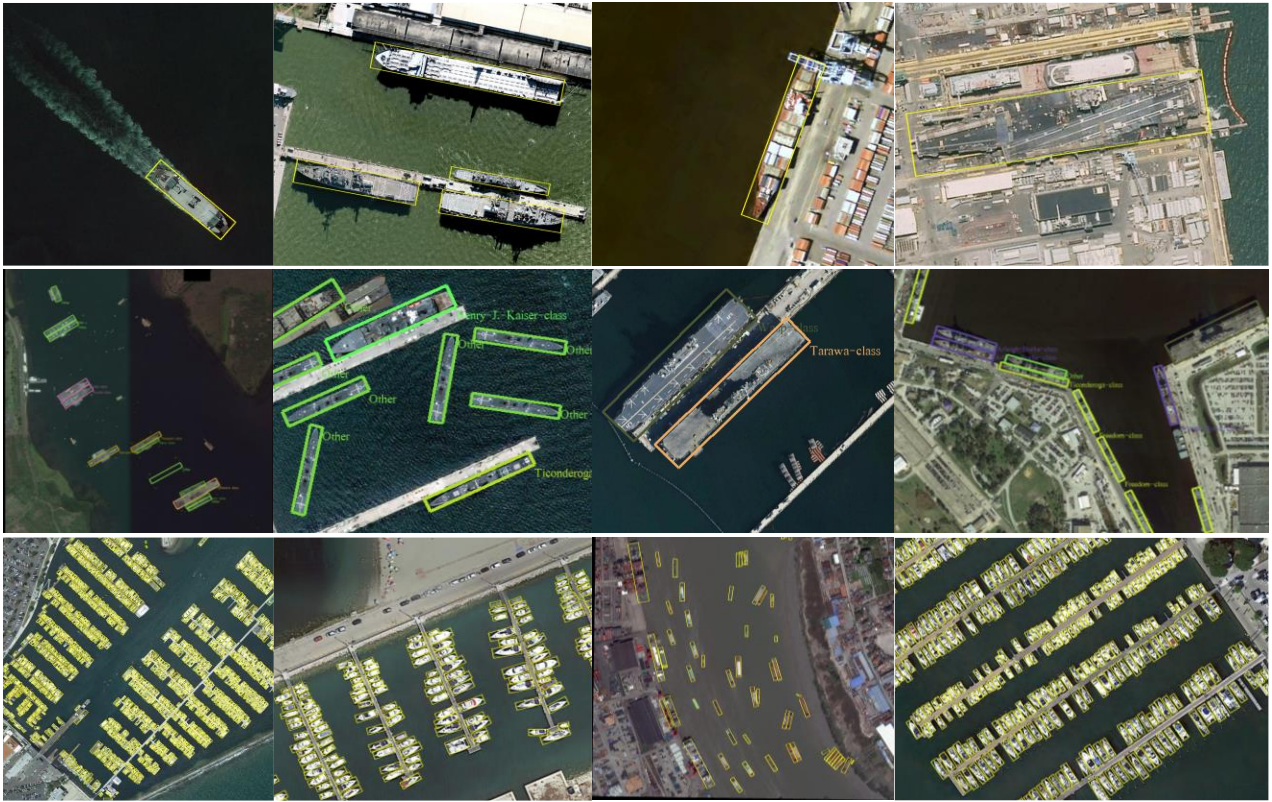

Figure 5. Visualization of the detection results of MidNet. The three rows are from HRSC2016, FGSD2021 and DOTA dataset respectively, where different colors are used in FGSD2021 to represent different categories.

Table 5. Ablation experiments for post-processing method on HRSC2016 dataset.

| Method | mAP |
|---|---|
| Keypoint matching based on ExtremeNet pipeline | 90.04 |
| Construct bounding box without gradient calculation | 90.45 |
| Our method | **90.52** |

## 5. Conclusion

We present a novel midpoints-based oriented ship detection framework. It predicts the position and matching vector of five keypoints (midpoints and center) and groups these keypoints to construct the oriented bounding box in a purely analytical geometry manner. To mining the discriminative features of keypoints in the complex background, a symmetrical deformable convolution is proposed based on the structural characteristics of ships. Our method provides a new perspective for predicting the oriented object by means of keypoints, which avoids unstable angle prediction and over-parameterized anchor setting. The state-of-the-art detection results on HRSC-2016 and FGSD2021 and competitive result on DOTA demonstrate the superiority of the proposed framework.

Meanwhile, we acknowledge that the generalization of our method in extremely irregular categories is limited, this is because the midpoints of these bounding boxes often seriously deviate from the objects. In the future work, we will try to explore weakly supervised learning methods and design adaptive feature learning methods to locate the keypoints of irregular objects precisely.